\documentclass[letterpaper, 10 pt, journal, twoside]{IEEEtran} 
\IEEEoverridecommandlockouts                              % This command is only
\usepackage{amsmath} % for \eqref
\usepackage{graphics} % for pdf, bitmapped graphics files
\usepackage{epsfig} % for postscript graphics files
\usepackage{mathptmx} % assumes new font selection scheme installed
\usepackage{times} % assumes new font selection scheme installed
\usepackage{amsmath} % assumes amsmath package installed
\usepackage{amssymb}  % assumes amsmath package installed
\usepackage{comment}
\usepackage{tikz}
\usepackage{cite}
\usepackage{import}
\graphicspath{{Images}}
\usepackage{color}

\usepackage{hyperref}
\hypersetup{%
  draft,
  breaklinks,
  bookmarksnumbered=true,
  bookmarksopen=true,
  colorlinks=true,
  linkcolor=black,
  urlcolor=black,
  citecolor=black
}

\title{
Contact-Implicit Trajectory Optimization using Orthogonal Collocation
}

\author{Amir Patel$^{1}$, Stacey Shield$^{1}$, Saif Kazi$^{2}$, Aaron M. Johnson$^{3}$, and Lorenz T. Biegler$^{2}$% <-this % stops a space
\thanks{Manuscript received: September, 10, 2018; Revised December, 12, 2018; Accepted January, 30, 2019.}%Use only for final RAL version
\thanks{This paper was recommended for publication by Editor Nancy Amato upon evaluation of the Associate Editor and Reviewers' comments. 
This work was  supported by  the National Research Foundation of South Africa (Grant: 99380) and the Oppenheimer Memorial Trust.}
\thanks{$^{1}$Amir Patel and Stacey Shield are with the Department of Electrical Engineering,  University of Cape Town, South Africa. 
    {\tt\footnotesize a.patel@uct.ac.za}}%
\thanks{$^{2}$ Saif Kazi and Lorenz T. Biegler are with the Department of Chemical Engineering, Carnegie Mellon University, USA.
    {\tt\footnotesize biegler@cmu.edu}}%
\thanks{$^{3}$Aaron M. Johnson is with the Department of Mechanical Engineering, Carnegie Mellon University, USA.
    {\tt\footnotesize amj1@cmu.edu}}%
\thanks{Digital Object Identifier (DOI): \href{https://doi.org/10.1109/LRA.2019.2900840}{10.1109/LRA.2019.2900840} }       
}

\newcommand\copyrighttext{%
 \textcopyright 2019 IEEE. Personal use of this material is permitted.
  Permission from IEEE must be obtained for all other uses, in any current or future
  media, including reprinting/republishing this material for advertising or promotional
  purposes, creating new collective works, for resale or redistribution to servers or
  lists, or reuse of any copyrighted component of this work in other works.
  DOI: \href{https://doi.org/10.1109/LRA.2019.2900840}{10.1109/LRA.2019.2900840}}
\newcommand\copyrightnotice{%
\begin{tikzpicture}[remember picture,overlay]
\node[anchor=south,yshift=5pt] at (current page.south) {\fbox{\parbox{\dimexpr\textwidth-\fboxsep-\fboxrule\relax}{  \footnotesize \copyrighttext}}};
\end{tikzpicture}%
}

\begin{document}

\bstctlcite{IEEEexample:BSTcontrol}

\thispagestyle{empty}
\setcounter{page}{0}
\begin{figure*}[t!]
\centering
\large
This paper has been accepted for publication in IEEE Robotics and Automation Letters.\\

DOI: \href{https://doi.org/10.1109/LRA.2019.2900840}{10.1109/LRA.2019.2900840}\\

IEEE Explore: \href{https://ieeexplore.ieee.org/document/8648229/}{https://ieeexplore.ieee.org/document/8648229/}\\

~\\

Please cite the paper as:\\

Amir Patel, Stacey Shield, Saif Kazi, Aaron M. Johnson, and Lorenz T. Biegler, ``Contact-Implicit Trajectory Optimization Using Orthogonal Collocation,'' in \emph{IEEE Robotics and Automation Letters}, vol. 4, no. 2, pp. 2242--2249, April 2019.\\

~\\

~\\

\copyrighttext
\vspace{400px}
\end{figure*}
\maketitle
\copyrightnotice
%\thispagestyle{empty}
%\pagestyle{empty}

%%%%%%%%%%%%%%%%%%%%%%%%%%%%%%%%%%%%%%%%%%%%%%%%%%%%%%%%%%%%%%%%%%%%%%%%%%%%%%%%
\begin{abstract}
In this paper we propose a method to improve the accuracy of trajectory optimization for dynamic robots with intermittent contact by using orthogonal collocation.  
Until recently, most trajectory optimization methods for systems with contacts employ mode-scheduling, which requires an a priori knowledge of the contact order and thus cannot produce complex or non-intuitive behaviors. Contact-implicit trajectory optimization methods offer a solution to this by allowing the optimization to make or break contacts as needed, but thus far have suffered from poor accuracy. Here, we combine methods from direct collocation using higher order orthogonal polynomials with contact-implicit optimization to generate trajectories with significantly improved accuracy. The key insight is to increase the order of the polynomial representation while maintaining the assumption that impact occurs over the duration of one finite element. 
\end{abstract}


%%%%%%%%%%%%%%%%%%%%%%%%%%%%%%%%%%%%%%%%%%%%%%%%%%%%%%%%%%%%%%%%%%%%%%%%%%%%%%%%
\section{INTRODUCTION}
\IEEEPARstart{C}{ontact} is ubiquitous in nature. Animals utilize contact with their environment to enable dexterous manipulation of objects or agile locomotion. In order for robots to one day perform similarly complex maneuvers in challenging environments, contacts must be adequately considered (and exploited) during motion planning. However, achieving this is still challenging as impulsive contact represents a significant numerical challenge for current methods.

One recent approach to solving this problem, introduced by Posa \cite{Posa2014AContact} and Mordatch \cite{mordatch2012discovery}, is \emph{contact-implicit optimization} (also called \emph{contact-invariant optimization} or \emph{through-contact optimization}). This method encodes both the contact force and the body state as part of an optimization problem, with the contact consistency enforced by algebraic constraints. 
Unlike hybrid systems models,
which use a fixed contact sequence defined a priori (or with an outer loop optimization) for the full system \cite{kelly2017introduction,hereid2018dynamic,posa2016optimization,wampler2009optimal,Pardo-RSS-17} or per-leg \cite{winkler2018gait},
or Mixed Integer Programming (MIP) \cite{aceituno2018simultaneous,kuindersma2016optimization},
which encodes the contact mode in an integer variable,
in contact-implicit optimization there is no variable specifying the contact mode at each time, and the contact is determined implicitly by the optimization. These methods are very effective for systems with many possible contacts or an unknown optimal contact sequence, and have enabled, e.g., discoveries of novel legged gaits \cite{xi2016selecting,yesilevskiy2018spine} and balancing trajectories \cite{shield2017balancing}.

\begin{figure}[tb]
	\centering
	\includegraphics[scale=0.6]{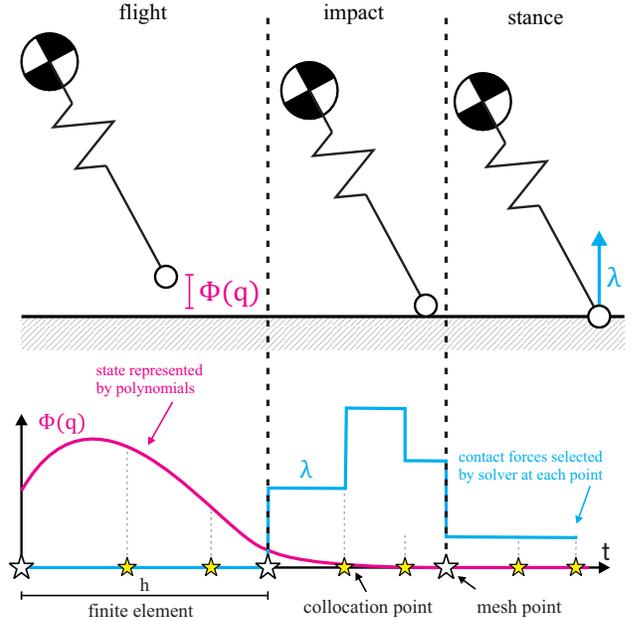}
	\caption{The concept of contact-implicit optimization using orthogonal collocation is shown. The state trajectory is expressed as a series of high-order polynomials on finite elements. Contact mode changes are enforced to only occur at the edges (mesh points) of the finite elements, ensuring smoothness and accuracy of the transcription.}
	\label{fig:overview} 
\end{figure}

However, most implementations only employ a first-order (Euler) integration method \cite{Posa2014AContact,mordatch2012discovery,shield2017balancing} or a pseudo-trapezoidal method \cite{xi2014optimal,xi2016selecting}, and are based on time-stepping simulation \cite{stewart2000implicit}. 
The disadvantage of first-order integration is that these methods requires a large number of steps, $N$, for sufficient accuracy (they have $O(1/N)$ accuracy), which limits their application to short time-horizon motions. 
These methods have been combined with variational integrators to produce a contact-implicit method with second-order accuracy \cite{Manchester2017}, as well as versions that use B-Splines \cite{werner2017generation} or Hermite-Simpson polynomials \cite{chao2017step}.
Previous works have noted that time-stepping methods can never exceed first-order discretization due to the fact that discontinuities could exist within the state trajectories \cite{anitescu2003fixed,anitescu2004constraint,acary2008numerical,xi2014optimal}. 

In this paper, we show that higher-order methods can indeed be employed and that doing so increases the accuracy of contact-implicit trajectory optimization. Our method uses higher-order orthogonal collocation (i.e.\ it solves the dynamics at collocation points based on $K\text{-th}$ degree orthogonal polynomials) \cite{hargraves1987direct,biegler2010nonlinear}
to provide a smoother representation of the state and velocity at every point on the interior of a finite element (for $O((1/N)^{2K-1})$ accuracy). 

It is this more accurate smooth representation that 
\cite{anitescu2003fixed,anitescu2004constraint,acary2008numerical,xi2014optimal} point to as the problem with higher-order methods, because a smooth representation would require an event-finding algorithm (i.e.\ collision detection) to isolate the non-smooth points. 
The presented method resolves this problem by approximating an event-driven simulation scheme (instead of time-stepping) within the context of a contact-implicit optimization problem.
That is, the presented method isolates the event times that the algebraic constraints are activated or deactivated and can apply an impulsive transition between contact modes. 
The key to this working is that we constrain the events (contact mode switches) to occur only at the mesh points (edges of the finite elements) as shown in Fig.~\ref{fig:overview}.
So long as the optimizer is given sufficient freedom to adjust the times that these mesh points represent, it implicitly solves the event-finding problem as part of the optimization. This eliminates the need for an additional collision detection algorithm. 
To make the problem formulation more tractable, we also present a relaxation that still maintains several key physical assumptions (especially impulses acting over the duration of one finite element) from the first-order implementations. We show that this relaxation approximates the exact formulation in the limit as time-steps are allowed to go to zero.

The paper is organized as follows: Sec.~\ref{sec:method} describes the method, starting with a background on collocation for dynamic (but smooth) trajectory optimization (Sec.~\ref{sec:collocation}) and then contact-implicit trajectory optimization (Sec.~\ref{sec:cio}). In Sec.~\ref{sec:ourapproach}, the application of orthogonal collocation to contact-implicit optimization problems is presented, with further implementation considerations presented in Sec.~\ref{sec:details}. Then, Sec.~\ref{sec:results} presents three case studies which utilize the method, ranging from a simple point particle to a bipedal robot. Lastly, Sec.~\ref{sec:conc} concludes the paper with a discussion of the implications of the results and proposed future work.

%%%%%%%%%%%%%%%%%%%%%%%%%%%%%%%%%%%%%%%%%%%%%%%%%%%%%%%%%%%%%%%%%%%%%%%%%%%%%%%%
\section{Method}
\label{sec:method}

This paper considers trajectory optimization, where the dynamics and control inputs are transcribed into a nonlinear program (NLP) that can be solved with numerical optimization algorithms. Trajectory optimization can be formulated either by \emph{direct} or \emph{indirect} methods. Here, we focus on direct methods as indirect methods have known disadvantages \cite{kelly2017introduction}. Comprehensive tutorials on trajectory optimization can be found in \cite{biegler2010nonlinear,kelly2017introduction}.

\subsection{Direct Collocation}

\label{sec:collocation}
Direct collocation formulates the trajectory optimization problem as an NLP without the need for forward integration (as in shooting methods) \cite{betts2010practical}. This is done by discretizing the trajectories (state and control) into $N$ time periods (finite elements) using polynomials. In our particular case the trajectories are represented using a Runge-Kutta basis with K-collocation points. The advantage of this representation, when compared to others such as B-splines \cite{werner2017generation}, is that most of the polynomial coefficients have the same variable bounds as the profiles themselves as well as considerably better numerical accuracy \cite[Ch.~8]{biegler2010nonlinear}. 

For example, consider the state variable $z$:
\begin{gather}
\frac{dz}{dt} = f(z(t),t), \qquad z(0) = z_{0}.
\label{eq:exDiffEqn}
\end{gather}
For time $t$ in finite element $i$, this yields the following Runge-Kutta basis representation of the state variable:
\begin{gather}
z(t) = z_{i,0} + h_{i}\sum_{j=1}^K\Omega_{j}(\tau)\dot{z}_{ij}, \quad t \in [t_i,t_{i-1}],
\label{eq:RKbasisEqn}
\end{gather}
where $z_{i,0}$ is a coefficient that represents the state variable at the beginning of element $i$, $\dot{z}_{ij}$ represents $\frac{dz(t_{ij})}{d\tau}$, $h_i$ is the length of the finite element, $\tau$ the relative time within that element, and $\Omega_{j}(\tau)$ is a polynomial of order $K$, satisfying:
\begin{gather}
\Omega_{j}(\tau) = \int_0^\tau \bar{l}(\tau^\prime)d\tau^\prime, \quad \tau \in [0,1],
\label{eq:OmegaEqn}
\end{gather}
where $\bar{l}(\tau^\prime)=\prod_{k=1,\neq j}^K\frac{(\tau^\prime-\tau_k^\prime)}{(\tau_j^\prime-\tau_k^\prime)}$.

Here, we employ $K$-point Radau collocation (a Gauss-Jacobi polynomial) to solve the differential equation at selected points in time. Radau collocation has many attractive features (stability, stiff decay) as well as having equivalent accuracy to Implicit Runge Kutta (IRK) integration, $O(h^{2K-1}$) \cite{biegler2010nonlinear}. Using this, the state variable at each collocation point $k$ of finite element $i$ is represented as:
\begin{gather}
z_{i,k} = z_{i,0} + h_i\sum_{j=1}^K\Omega_j(\tau_k)\dot{z}_{ij}, \quad k\in \{1,\cdots,K\},
\label{eq:stateCollocationEqn}
\end{gather}
with $\Omega$ derived using (\ref{eq:OmegaEqn}) and $\tau_k$ the relative time of collocation point $k$. For a given $K$, values for the weightings $\Omega$ and time divisions $\tau$ can be found in \cite[Ch.~8]{biegler2010nonlinear}. Continuity at the finite element boundaries is enforced by:
\begin{gather}
z_{i,0} = z_{i-1,K}, \quad i\in\{2,\cdots,N\}.
\label{eq:stateContEqn}
\end{gather}

\subsection{Contact-Implicit Trajectory Optimization}
\label{sec:cio}
For contact-implicit trajectory optimization, the dynamics \eqref{eq:exDiffEqn} can be modeled as a rigid multi-body system using Euler-Lagrange mechanics with generalized coordinates $\mathbf{q}$ and often expressed in the form:
\begin{gather}
\mathbf{M}\mathbf{\ddot{q}} + \mathbf{C}\mathbf{\dot{q}} + \mathbf{G} = \mathbf{B}\mathbf{u}+\mathbf{J^T}\mathbf{\lambda},
\label{eq:manipulatorEqn}
\end{gather}
where $\mathbf{M}$ represents the mass matrix, $\mathbf{C}$ the Coriolis and centrifugal matrix, $\mathbf{G}$ the gravitational force, $\mathbf{B}$ the input mapping, $\mathbf{u}$ the generalized input, $\mathbf{J}$ the contact Jacobian, and $\mathbf{\lambda}$ the contact forces. 
Multi-body systems with contact possess hybrid dynamics, where $\mathbf{\lambda}$ only acts in specific configurations of the state space (i.e.\ when in contact). 
When switching between contact conditions, Newtonian plastic impact says that an impulse $\Lambda$ at the contact location leads to a discontinuity in velocity from $\mathbf{\dot{q}}^-$ (pre-impact) to $\mathbf{\dot{q}}^+$ (post-impact), defined by\footnote{In closed form, these constraints define $\Lambda = -(\mathbf{J} \mathbf{M}^{-1} \mathbf{J}^T)^{-1} \mathbf{J} \mathbf{\dot{q}}^-$, as in e.g.~\cite[Eqn.~25]{paper:johnson_hs_2016}, however in this optimization context it is more convenient to leave the definition of the impulse $\Lambda$ implicit. While other models of impact dynamics could be used, we believe that plastic impact is the most appropriate model for robot dynamics for the reasons given in \cite[A8]{paper:johnson_hs_2016}.}:
\begin{gather}
\mathbf{M}(\mathbf{\dot{q}}^+ - \mathbf{\dot{q}}^-) = \mathbf{J}^T \Lambda, \qquad \mathbf{J}\mathbf{\dot{q}}^+ = 0 \label{eq:impactEqn}
\end{gather}
which may be derived by taking the limit of a an impact event as the time duration goes to zero in \eqref{eq:manipulatorEqn}. 

Prior contact-implicit methods of trajectory optimization for these systems \cite{Posa2014AContact,mordatch2012discovery} are based off the time-stepping simulation method \cite{stewart2000implicit} which discretizes time and considers the combination of both forces $\lambda_i$ and impulses $\Lambda_i$ integrated over time-step $i$ in a combined $\lambda_i$. 
The dynamics are encoded using direct transcription using 1-point collocation, i.e.\ a first order approximation defined entirely by the value of the states and contact forces at each mesh point (the start of each finite element). By including $\lambda$ as part of the decision variables, the optimization implicitly chooses the sequence of contacts, represented as time-steps $i$ where $\lambda_i>0$.
The contact force requires the complementarity constraint:
\begin{gather}
\lambda_i^T\phi(\mathbf{q}_{i+1})=0, \quad \phi(\mathbf{q}_i)\geq 0, \quad \lambda_i\geq 0
\label{eq:complementEqn1}
\end{gather}
where $\phi(\mathbf{q})$ represents the non-penetration constraint between rigid bodies (and $\mathbf{J}= \frac{\partial \mathbf{\phi}}{\partial \mathbf{q}} $). Note that this formulation is not meant for systems with a large number of bodies or complex surfaces (e.g.\ billiards or walking on gravel) and assumes a finite number of contact constraints.
The choice of indices is important to ensure that a contact force or impulse over a time-step $i$ enforces the equality $\phi_{i+1}= 0$ at the end of that time-step. Additional constraints enforce the friction cone\footnote{For clarity, we've only included described the 2D case, but these can easily be extended to 3D \cite{Manchester2017}.}~\cite{Posa2014AContact}:
\begin{align}
\lambda_{y,i}\geq 0, \quad \lambda_{x,i}^+\geq 0, \quad \lambda_{x,i}^-&\geq 0,
\label{eq:complementEqn2}\\
\mu\lambda_{y,i}-\lambda_{x,i}^+-\lambda_{x,i}^-&\geq0,
\label{eq:complementEqn3}\\
(\mu\lambda_{y,i}-\lambda_{x,i}^+-\lambda_{x,i}^-)^T\gamma_i&=0,
\label{eq:complementEqn4}
\end{align}
with $\mathbf{\lambda}=[\lambda_{x,i}^+-\lambda_{x,i}^-\,,\; \lambda_{y,i}]^T$ where $x$ is the direction tangent to the contact surface and $y$ is the direction normal to it, while $\gamma_i$ is the magnitude of the relative tangential velocity at the point of contact. Additionally, if the contact point is sliding, it is constrained to do so with a frictional force along the edge of the friction cone:
\begin{align}
\gamma_i+\psi(q_i,\dot{q_i})&\geq 0
\label{eq:complementEqn5}\\
\gamma_i-\psi(q_i,\dot{q_i})&\geq 0
\label{eq:complementEqn6}\\
\lambda_{x,i}^{+T}(\gamma_i+\psi(q_i,\dot{q_i}))&=0
\label{eq:complementEqn7}\\
\lambda_{x,i}^{-T}(\gamma_i-\psi(q_i,\dot{q_i}))&=0
\label{eq:complementEqn8}
\end{align}
with $\psi(q_i,\dot{q_i})$ the relative tangential velocity at the contact.

The constraints (\ref{eq:complementEqn1}), (\ref{eq:complementEqn4}), (\ref{eq:complementEqn7}), (\ref{eq:complementEqn8}) transform the NLP into a Mathematical Program with Equality Constraints (MPEC) which is notoriously difficult to solve. The two main methods for making the MPEC problem more tractable are the $\epsilon$-relaxation method and the penalty method
%, and the slack variable method
\cite{ralph2004some,hoheisel2013theoretical,fletcher2004solving}. In the $\epsilon$-relaxation method, the complementarity constraints are reformulated as a set of inequality constraints, relaxed by a parameter $\epsilon>0$:
\begin{gather}
\alpha^T\beta = 0 \; \Rightarrow \; \alpha^T\beta\leq \epsilon,
\label{eq:relaxEqn}
\end{gather}
where $\alpha$ and $\beta$ are positive slack variables. The MPEC is then solved as series relaxed problems decreasing $\epsilon$ to a user-defined accuracy. In the penalty method\cite{BAUMRUCKER20091248}, the complementarity constraint is removed and its $l_1$ norm is included in the objective:
\begin{eqnarray}
\min_z & {g(z)} + \rho\alpha^{T}\beta.
\label{eq:Prho}
\end{eqnarray} 
This allows the problem to appear more feasible to the NLP, but requires $\rho$ to be greater than some critical value $\rho_c$ in order to exactly satisfy the complementarity at the solution. 

In the examples described in Section~III, we find that the best MPEC solution strategy is problem dependent.

%%%%%%%%%%%%%%%%%%%%%%%%%%%%%%%%%%%%%%%%%%%%%%%%%%%%%%%%%
\subsection{Our Approach}
\label{sec:ourapproach}

\begin{figure}[tb]
	\centering
	\vspace{.3em}
	\includegraphics[scale=0.6]{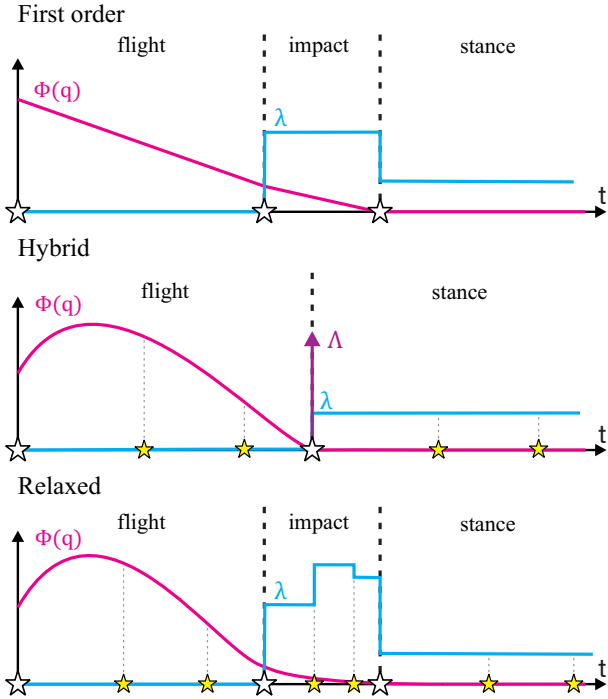}
	\caption{Comparison of different approaches to handling impact. In first order methods, the state is piecewise-linear and the effect of contact forces and impulses are combined over a finite element. In the hybrid-system formulation, the state is a smooth polynomial within a finite element and the optimization solves for the contact forces, impulses, and times. In the relaxed formulation, the smooth state representation is maintained everywhere but the contact impulse is again combined with contact force over a finite element. The impact element is exaggerated in length for clarity.}
	\label{fig:impact_comparison_fig} 
\end{figure}

In this paper, we use direct collocation with higher-order orthogonal polynomials to represent the state $\mathbf{q}$ and velocity $\mathbf{\dot{q}}$ of a
contact-implicit optimization, while enforcing the dynamics at both the mesh points and the collocation points. 
Prior first-order methods evaluate the dynamics at each mesh point and use a linear interpolation in between (Fig.~\ref{fig:impact_comparison_fig}). These first-order methods do not determine where within a finite element an impact occurs, because the effect of an impulse $\Lambda$ is spread out over the entire element. 
However, only having a linear interpolation limits their accuracy both by not locating the point of impact as well as not following other nonlinear changes in the trajectory. 

Here, with a higher-order representation of the system within a finite element, contact changes are constrained to only occur at the mesh points. If this is not done, as in \cite{chao2017step,werner2017generation}, complementarity is not guaranteed within the finite element (e.g.\ the foot could leave or strike the ground at a collocation point). 
Using a higher-order representation with contact changes at the mesh points  increases the accuracy of the trajectory both by isolating the time of impacts and by tracking the continuous dynamics more closely. 

By isolating the event times, the optimization can be considered as approximating an event-driven simulation scheme instead of a time-stepping scheme. In event-driven simulation, the continuous dynamics are integrated with a numerical ODE (ordinary differential equation) or DAE (differential-algebraic equation) integration algorithm (e.g.\ \texttt{ode45}) and stopped when an event is detected to handle the impulsive change in contact conditions. A time-stepping scheme combines the effect of continuous dynamics over a fixed time-step with the impulsive contact changes in a single step, performing a first-order integration of the system as an MDI (measure differential inclusion). 
These are typically considered separate classes of numerical algorithms (e.g.~\cite[Sec.~6.3]{brogliato2002numerical}), but in this setting lying along the same continuum. 

Once the events are constrained to the mesh points, the question then becomes how to handle the discontinuous dynamics of impact, \eqref{eq:impactEqn}. This was not an issue in the first-order methods since they are already non-smooth everywhere. Here we present two possible solutions: the full hybrid-system formulation and a more tractable relaxed approximation. 

\paragraph{Hybrid-system Formulation} The hybrid-system dynamics of \eqref{eq:impactEqn} can be explicitly included in the optimization by replacing the velocity continuity equations \eqref{eq:stateContEqn}. Contact impulses $\Lambda_i$ at each mesh point $i$ are added to the set of decision variables (Fig.~\ref{fig:impact_comparison_fig}), just as contact forces $\lambda_i$ are in any contact-implicit scheme, with the additional constraints:
\begin{gather}
\Lambda_i^T\phi(\mathbf{q}_{i+1})=0, \quad \phi(\mathbf{q}_i)\geq 0, \quad \Lambda_i\geq 0
\label{eq:ImpulseComplementEqn1}
\end{gather}
as well as constraints analogous to \eqref{eq:complementEqn2}--\eqref{eq:complementEqn4} in the case of frictional impact. 

\paragraph{Relaxed Formulation} However, this hybrid formulation introduces even more complementarity conditions and thus far has only resulted in a solvable optimization problem when initialized very close to the correct solution. Therefore, we relax the hard-impact constraint by spreading the impulse out over the duration of a finite element (Fig.~\ref{fig:impact_comparison_fig}). This may at first seem to violate the rigid-body model of physics, as the impulse start to act before the object reaches the contact. However, this is the same relaxation that the first-order methods use implicitly \cite{Posa2014AContact,mordatch2012discovery}, the only difference is that the higher-order version approximates the intermediate trajectory smoothly while the first-order case does not. The other contact constraints still hold -- the impulse may act just before contact is made only if the bodies do in fact make contact at the end of that finite element.

The impulse (encoded as part of the contact force) is only allowed to be applied for the duration of one finite element. That time, $h_i$, is a decision variable. In the limit, if we let $h_i$ become very small, the equations of motion converge to exactly the plastic impact law of \eqref{eq:impactEqn}, as seen in Sec.~\ref{sec:ball} and Fig.~\ref{fig:ballVelFig}. In practice, the optimization algorithm naturally chooses small $h_i$ with comparatively large contact forces as it uses this freedom to approximate the rigid-body impact.

Using the complementarity formulation of \eqref{eq:complementEqn1}, requiring contact over finite element $i+1$ to enable contact forces over finite element $i$, also produces a challenging constraint at liftoff. Just as the contact force could act before touchdown, it must also cease before liftoff. Smooth (nonimpulsive) liftoff occurs when the contact force goes to zero anyway, so this constraint is not as tricky as the touchdown constraint. However, the optimizer must have the freedom to either use sufficient control effort to maintain zero contact force or reduce the time duration of the liftoff event to a small value of $h_i$. This also implies that contacts must persist for a minimum dwell time of at least one finite element, precluding exact solutions to simultaneous but sequential transitions (e.g.~\cite[Thm.~8]{paper:johnson_hs_2016}). An example showing this limitation is given in Sec.~\ref{sec:ball}.

\subsection{Implementation Details}
\label{sec:details}
A previous application of orthogonal collocation to MPEC optimizations by Baumrucker \& Biegler \cite{BAUMRUCKER20091248} enforces mode changes at the mesh points by complementing one  variable with the $L_1$ norm of the other within the finite element. A new slack variable is introduced:
\begin{gather}
\alpha^\prime_{i}= \sum_{j=0}^K\alpha_{ij},
\label{eq:baumruckerEqn1}
\end{gather} 
and then the complementarity is expressed as:
\begin{gather}
\alpha^\prime_{i}\beta_{ij}=0.
\label{eq:baumruckerEqn2}
\end{gather} 
This solves the complementarity constraint at each collocation point and ensures that the mode is constant within the finite element. This formulation ensures accuracy but results in $N$ x $K$ complementarity equations.

To apply this to contact-implicit optimization, we propose an additional set of slack variables:
\begin{gather}
\beta^\prime_{i}=\sum_{j=0}^K\beta_{ij},
\label{eq:OurCompEqn1}
\end{gather} 
with the following reformulated complementarity:
\begin{gather}
\alpha^\prime_{i}\beta^\prime_{i}=0,
\label{eq:OurCompEqn2}
\end{gather} 
which increases the problem size, but results in the complementarity constraints only being evaluated once at each mesh point, while ensuring contact mode is fixed within the finite element. This formulation can readily be applied to the complementarity equations \eqref{eq:complementEqn1}, \eqref{eq:complementEqn4}, \eqref{eq:complementEqn7} and \eqref{eq:complementEqn8}.

An additional change that must be introduced when increasing the number of collocation points is that the control input, $\mathbf{u}$, must be constrained within a finite element. 
While the control variables could be represented using Lagrange polynomials, with discontinuities at the mesh points, in practise this leads to an artificial oscillation in control signal over the course of a finite element which leads to slow convergence (see \cite{biegler2010nonlinear} for discussion). Here, we employ a piecewise-constant control within the finite element as this mitigates the problem of singular arcs in the control problem (though other more refined control laws may be used instead). An example of why this is necessary is given in Sec.~\ref{sec:biped} and Fig.~\ref{fig:hipTorqComp_fig}.

In summary, the optimization problem has the following decision variables, indexed by system state $h$, finite element $i$, collocation point $j$, contact point $k$, control input $l$, and direction $m \in \{x,y\}$:\\[0.3em]
\begin{tabular}{l l}
$q_{h,i,j}$ & System state\\
$\dot{q}_{h,i,j}$ & System velocity\\
$\ddot{q}_{h,i,j}$ & System acceleration\\
$\lambda_{y,i,j,k}$ & Normal contact force \\
$\lambda_{x,i,j,k}^+$ & Tangential positive contact force \\
$\lambda_{x,i,j,k}^-$ & Tangential negative contact force \\
$u_{i,l}$ & Control input\\
$h_i$ & Time duration\\
$\alpha_{m,i,j},\alpha_{m,i}^\prime$ & slack variables for $\phi_{i,j,k}$ or $\gamma_{i,j,k}$\\
$\beta_{m,i,j},\beta_{m,i}^\prime$ &  slack variables for $\lambda_{m,i,k}$\\
\end{tabular}\\[0.5em]
For $H$ system states, $N$ finite elements, $K$ point collocation, and $C$ contact points, and $U$ control inputs, this results in $N(K(3H+3C)+U+1+8C(K+1))$ decision variables (for planar friction). 

The optimization problem has the following constraints:
\\[0.3em]
\begin{tabular}{l l}
\eqref{eq:manipulatorEqn} & Acceleration dynamics \\ %NKH
\eqref{eq:stateCollocationEqn}, \eqref{eq:stateContEqn} & Collocation constraints for $q$, $\dot{q}$ \\ % $2NKH$ Collocation (K-1) + continuity 1
\eqref{eq:complementEqn1} & Normal complementarity  \\ %$3NC$
\eqref{eq:complementEqn2} -- \eqref{eq:complementEqn8} & Frictional complementarity\\  %  $9NC$
\eqref{eq:baumruckerEqn1}, \eqref{eq:OurCompEqn1} & Slack variable definitions and constraints \\ % $4*2*N*C$ for primes, $ 4*2*K*N*C$ for ijs
\end{tabular}\\[0.5em]
for a total of $N(3KH + C(20+8K))$ constraints, in addition to problem-specific constraints such as initial and final conditions, bounds on variables such as $h_i$, or input constraints. 
The objective function, $g$, is also problem-specific, with minimum-time, minimum-effort, and similar functions commonly used.
This results in a large but sparse optimization problem which is particularly suited for sparse nonlinear solvers such as IPOPT \cite{wachter2006implementation}, CONOPT \cite{drud1995system}, or SNOPT \cite{gill2005snopt}. For the results in this paper, the optimization problems were written in GAMS \cite{bussieck2004general} and solved with either IPOPT or CONOPT. This code is available  online:\\
{\footnotesize \href{https://github.com/UCTMechatronics/orthogonal-collocation-with-contacts}{https://github.com/UCTMechatronics/orthogonal-collocation-with-contacts}}

%%%%%%%%%%%%%%%%%%%%%%%%%%%%%%%%%%%%%%%%%%%%%%%%%%%%%%%%%%%%%%%%%%%%%%%%%%%%%%%%
\section{Results}
\label{sec:results}

We implement the contact-implicit trajectory optimization with orthogonal collocation method on three example problems, with a focus on testing the hypothesis that it provides better accuracy. We also show the implications of some of the formulation decisions discussed in Sec.~\ref{sec:ourapproach} and~\ref{sec:details}.

\subsection{Ball Hitting Ceiling} 
\label{sec:ball}
In this example, we simulate the trajectory of a ball (point mass) colliding with a ceiling. The ball has an initial upward velocity and is acted on by gravity and a contact force when colliding with the ceiling. This example was chosen to demonstrate the ability to successfully capture impulse-like behavior as the lower bound on finite element time, $h_L \leq h_i$, is reduced. The system was implemented using 3-point Radau collocation, with total simulation time ($T$) of 1s and 100 finite elements. The optimization was performed in GAMS using the CONOPT solver and the complementarity constraints were formulated using the penalty method. To assess accuracy, a hybrid-dynamic simulation was also implemented in Matlab using the \texttt{ode45} solver (accuracy 1e-12).

\begin{figure}[tb]
	\centering
    \vspace{.5em}
	\def\svgwidth{\columnwidth}
    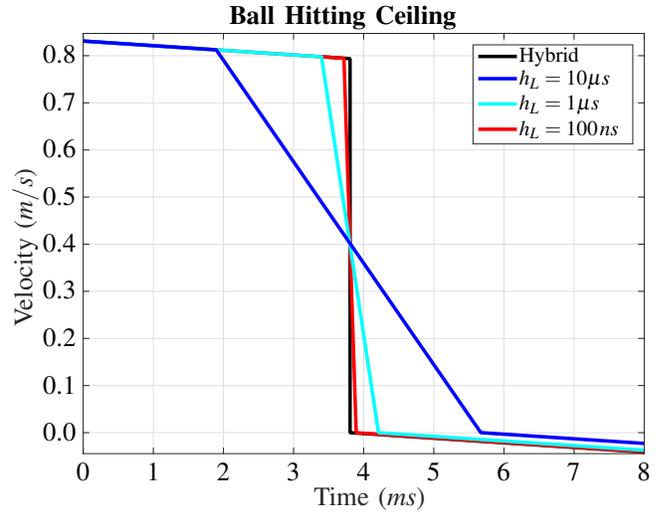
	\caption{Velocity at the point of impact is depicted. As the time-step lower bound ($h_L$) is decreased, the solution approaches the true, discontinuous solution. %The RMS errors for each of the trajectories are 0.0948, 0.0374, 0.0167 and 0.0067 respectively.
	}
	\label{fig:ballVelFig}
\end{figure}

The results show that lowering the time-step bound increases the accuracy of the solution and in the limit the trajectory approaches the hybrid system trajectory, as seen in Fig.~\ref{fig:ballVelFig}. Even in the worst case here, the impact event duration is around $2.5ms$ (and the other examples are all less than $1ms$). The system is able to capture the sequential-but-simultaneous impact and liftoff transitions \cite[Thm.~8]{paper:johnson_hs_2016}, lasting one finite element each, however the liftoff is slightly delayed compared to the hybrid solution.

\subsection{Double Pendulum with Hard-Stops Swing-Up} 
\label{sec:swingup}

\begin{figure}[tb]
	\centering
	\vspace{.6em}
	\def\svgwidth{0.75\columnwidth}
%% Accompanies image file 'pendulum_diagram_latex.eps' (pdf, eps, ps)
\begingroup%
  \makeatletter%
  \providecommand\color[2][]{%
    \errmessage{(Inkscape) Color is used for the text in Inkscape, but the package 'color.sty' is not loaded}%
    \renewcommand\color[2][]{}%
  }%
  \providecommand\transparent[1]{%
    \errmessage{(Inkscape) Transparency is used (non-zero) for the text in Inkscape, but the package 'transparent.sty' is not loaded}%
    \renewcommand\transparent[1]{}%
  }%
  \providecommand\rotatebox[2]{#2}%
  \newcommand*\fsize{\dimexpr\f@size pt\relax}%
  \newcommand*\lineheight[1]{\fontsize{\fsize}{#1\fsize}\selectfont}%
  \ifx\svgwidth\undefined%
    \setlength{\unitlength}{302.11748985bp}%
    \ifx\svgscale\undefined%
      \relax%
    \else%
      \setlength{\unitlength}{\unitlength * \real{\svgscale}}%
    \fi%
  \else%
    \setlength{\unitlength}{\svgwidth}%
  \fi%
  \global\let\svgwidth\undefined%
  \global\let\svgscale\undefined%
  \makeatother%
  \begin{picture}(1,1.0037864)%
    \lineheight{1}%
    \setlength\tabcolsep{0pt}%
    \put(0,0){\includegraphics[width=\unitlength]{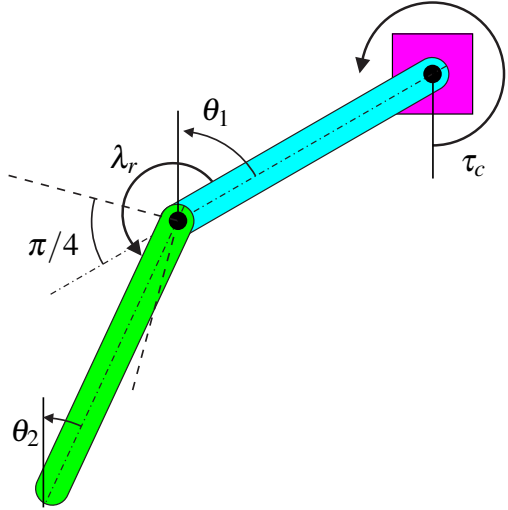}}%
    \put(0.00651165,0.1347218){\color[rgb]{0,0,0}\makebox(0,0)[lt]{\lineheight{1.25}\smash{\begin{tabular}[t]{l}\large $\theta_2$\end{tabular}}}}%
    \put(0.20524384,0.68186019){\color[rgb]{0,0,0}\makebox(0,0)[lt]{\lineheight{1.25}\smash{\begin{tabular}[t]{l}\large $\lambda_r$\end{tabular}}}}%
    \put(0.03961054,0.50817296){\color[rgb]{0,0,0}\makebox(0,0)[lt]{\lineheight{1.25}\smash{\begin{tabular}[t]{l}\large $\pi/4$\end{tabular}}}}%
    \put(0.38846898,0.77752969){\color[rgb]{0,0,0}\makebox(0,0)[lt]{\lineheight{1.25}\smash{\begin{tabular}[t]{l}\large $\theta_1$\end{tabular}}}}%
    \put(0.90693341,0.6754827){\color[rgb]{0,0,0}\makebox(0,0)[lt]{\lineheight{1.25}\smash{\begin{tabular}[t]{l}\large $\tau_c$\end{tabular}}}}%
  \end{picture}%
\endgroup%

	\caption{Double pendulum model with hard stops at the center joint.}
	\label{fig:pendulum_diagram}
\end{figure}

The second test system is a double pendulum with hard stops in the center joint, Fig.~\ref{fig:pendulum_diagram}. For this system we compare the solving time and accuracy of three- and five-point Radau collocation algorithms (R3 and R5) to implicit Euler (IE) \cite{Posa2014AContact} and variational integration (VI) \cite{Manchester2017}.  Using a torque $\tau_c$ acting at the base of the first link, the pendulum must swing itself up to a stationary vertical configuration. 
 
The second link is constrained with hard contacts to swing only within $\pi/4$ radians relative to the first link. The following constraints assure that the rebound torque $\lambda_r$ can only act when the link hits these bounds:
\begin{gather}
\alpha_{up}^+, \alpha_{up}^-, \alpha_{lo}^+, \alpha_{lo}^-, \lambda_{r}^+, \lambda_{r}^- \geq 0
\label{eq:pedulum_slackvars}\\
\alpha_{up}^+-\alpha_{up}^- = \frac{\pi}{4} - (\theta_2-\theta_1)
\label{eq:pendulum_upperbound}\\
\alpha_{lo}^+-\alpha_{lo}^- = -\frac{\pi}{4} - (\theta_2-\theta_1)
\label{eq:pendulum_lowerbound}\\
\alpha_{up}^+\lambda_{r}^-=0
\label{eq:pendulum_comp1}\\
\alpha_{lo}^-\lambda_{r}^+=0
\label{eq:pendulum_comp2}
\end{gather}

Solving this trajectory optimization problem is challenging for most NLP solvers. One strategy for improving the convergence rate is to provide a feasible (or close to feasible) initial solution \cite{safdarnejad2015initialization}. As such, this problem was solved in two sequential stages: first solving for a feasible problem (no cost function) and then using this solution as a seed to the full problem to optimize the cost function,
\begin{gather}
g(z) = \sum_{i=1}^N \tau_{c,i}^2h_i,
\label{eq:pendulum_cost}
\end{gather}
%with $P$ fixed to zero. 
The active set solver CONOPT was utilized and the penalty method~\eqref{eq:Prho} was used to formulate the complementarity problem. The state variables were initialized with random values between $-\pi$ and $\pi$, while all other variables were set to a fixed nonzero value (0.01). Each algorithm was tested with 600 random seeds for three different problem sizes: $N$=50, 100, and 200 elements. 
Further tests using 600 and 1000 elements were run in the case of the IE algorithm, to correspond to the combined number of collocation points ($N \times K$) for $N$=200 with R3 and R5 (since the dynamics are evaluated at $N \times K$ points). 
The maneuver was executed in approximately 2 seconds, with the $h_i$ allowed to vary within $\pm 20$ percent of $2/N$ seconds. 

The accuracy of the solutions was tested by comparing the state at each collocation point to the value generated by integrating the dynamic equations from the start of the finite element using MATLAB's \texttt{ode45} solver with 1e-12 accuracy  \cite{BAUMRUCKER20091248}. Since the VI method does not explicitly use instantaneous velocities, these values were approximated by calculating the generalized momentum $p$ from the discrete Lagrangian $L_d$ as in \cite{stern2006discrete}:
\begin{gather}
p_i = D^2L_d(q_{i-1},q_i) + \tau_{c,i} + \tau_{r,i},
\label{eq:generalized_momentum}
\end{gather}
where, $D^2$ is the second derivative with respect to time. The velocities can then be evaluated using $p = \frac{\delta L}{\delta \dot{q}}$.

\begin{figure} [tb]
	\centering
	\vspace{.5em}
	\def\svgwidth{1.0\columnwidth}
    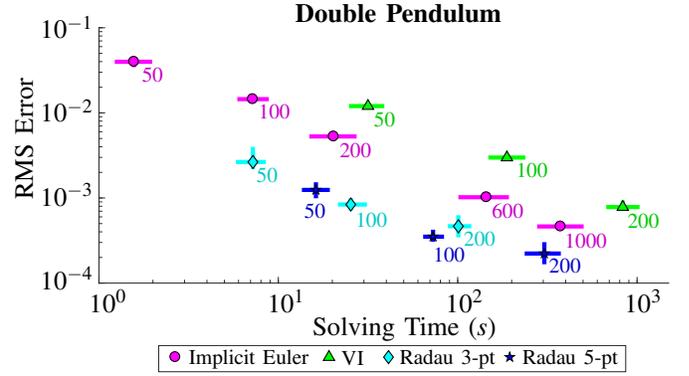
	\caption{Error and solving times for pendulum swing-up trajectories in all data sets. Bars indicate the inter-quartile range for 600 trials. Note that  extra tests for $N$=600 and 1000 for implicit Euler are still not as accurate as $N$=100 Radau 5-point and are also much slower.}
	\label{fig:pendulum_bar} 
\end{figure}

Each trial was run using four cores on a 32 core PC (Intel Xeon 2.2 GHz, 32 GB RAM). The median values and inter-quartile range for error and solving time for all problem are shown in Fig.~\ref{fig:pendulum_bar}. 
These results show that the IE and VI algorithms require more elements and longer solving times to achieve comparable accuracy to the Radau method (however see Sec.~\ref{sec:conc} for a discussion of when a variational integrator may be warranted). For example, note that the $N$=100 R5 method is both faster and more accurate than either $N$=600 or $N$=1000 for IE (both of which evaluate the dynamics at more points than the $N\times K$=500 points with R5).

\subsection{Biped}
\label{sec:biped}
One of the primary motivations for this work is to generate long-time-horizon optimal motions with legged robots. Here, we optimize a biped model (based on \cite{schultz2010modeling}) to perform a $10m$ run. However, we do not prescribe periodicity nor contact order. In addition to the contact points at the toes, we also include contact points at the heels, for 4 total contact points. The model is constrained to start and end standing upright and at rest with the terminal condition being $x(t_f)=10 m$ with the objective function of energy minimization as in  (\ref{eq:pendulum_cost}). A 1 m gap is also included, which the robot is required to negotiate.

\begin{figure*}
	\centering
	\vspace{.1em}
	\includegraphics[width=0.95\textwidth]{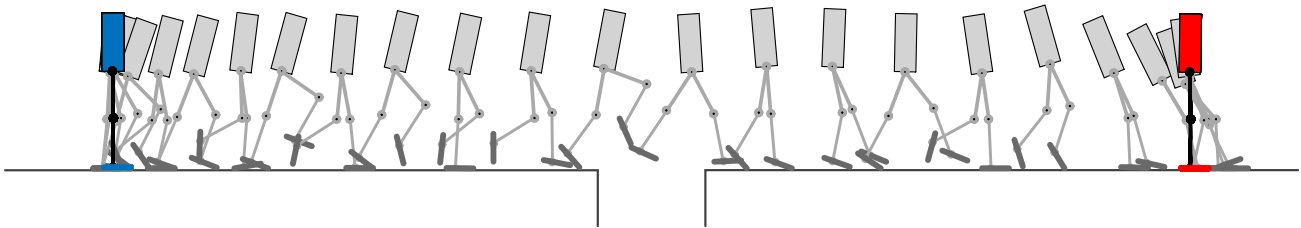}
	\caption{Animation of the resulting trajectory showing that the model starts and ends at rest while traversing 10m and negotiating a gap in the floor.}
	\label{fig:biped_animation} 
\end{figure*} 

The problem is discretized using $N=300$ elements, $K=3$ Radau collocation, and solved using IPOPT in GAMS. The complementarity constraints were formulated using the $\epsilon$-relaxation method. Due to the four contact points and long-time horizon, this problem is significantly more challenging to solve the previous examples having around 90,000 variables and 100,000 constraints. We employ two stages of initialization: First, the problem of finding a feasible (fixed cost) and $\epsilon$-relaxed ($\epsilon =10$) trajectory is solved with fixed time-steps. This trajectory is then used to seed the full problem, where the $h_i$ are allowed to vary within $\pm 50$ percent of $T/N$. Total convergence time was about 3 hours on the same PC described in \ref{sec:swingup}.

The resulting solution (Fig.~\ref{fig:biped_animation}) demonstrates that the method is able to generate motion plans without prescribing contact sequences for a multi-contact model. For comparison, an implicit Euler trajectory (see supplementary video) was also generated, however the resulting motion is quite unnatural, exhibiting excessive chattering and floating behaviour. As with the previous example, implicit Euler and 3-point Radau collocation were compared using \texttt{ode45} in Matlab and the higher order collocation improved the RMS error from $1.12\times10^{-2}$ to $2.04\times10^{-4}$.

\begin{figure} [tb]
	\centering
	\def\svgwidth{1.0\columnwidth}
    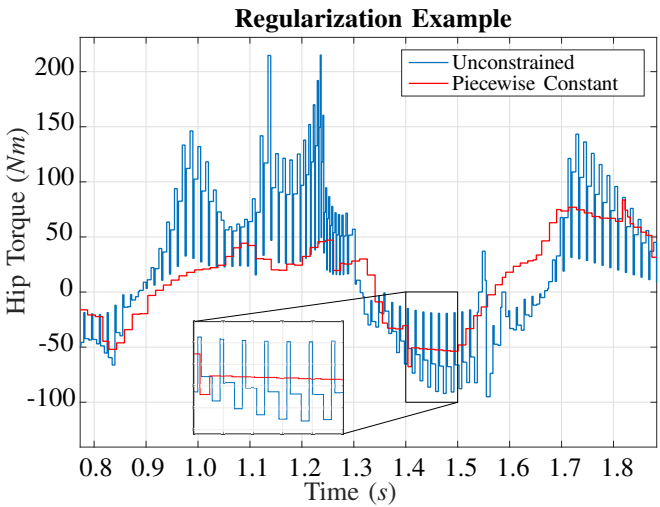
	\caption{A comparison between the unconstrained (changing at each collocation point) and piecewise constant (fixed over an entire finite element) torque is depicted. This illustrates the need for regularization of the control input to prevent oscillation and aid convergence.}
	\label{fig:hipTorqComp_fig} 
\end{figure}

To illustrate the need for input torque regularization, we re-ran the same optimization and allowed the input torque to be unconstrained within the finite element (i.e.\ vary at each collocation point within the element). This nearly doubled the convergence time over the regularized (piecewise-constant over the finite element) torque version. Additionally, as is evident by Fig.~\ref{fig:hipTorqComp_fig}, without regularization the torque oscillates, which would be undesirable when implementing this trajectory on a robot.

%%%%%%%%%%%%%%%%%%%%%%%%%%%%%%%%%%%%%%%%%%%%%%%%%%%%%%%%%%%%%%%%%%%%%%%%%%%%%%%%
\section{Discussion \& Future Work}
\label{sec:conc}

This paper presents a method for contact-implicit trajectory optimization which utilizes higher order orthogonal collocation to obtain a more accurate representation of the dynamics than previous methods. 
To avoid the problems of discontinuities in the smooth representation \cite{anitescu2003fixed,anitescu2004constraint,acary2008numerical,xi2014optimal}, we enforce mode changes to only occur at the mesh points. In contrast to the limitations of time-stepping simulation, where ``if no collision detection is performed, the integration method cannot exceed order one,'' \cite{xi2014optimal}, in the context of trajectory optimization where the time-steps are not fixed, collision detection is in fact being performed by the optimizer, allowing for higher order integration that approximates an event-driven simulation.

Transitioning from time-stepping to event-driven as the underlying simulation for the contact-implicit trajectory optimization is well justified in many robotic systems where the frequency of contact changes is relatively low (as opposed to, e.g., a billiard simulation). Because of this, the higher-order collocation allows for fewer finite elements to be utilized without loss of accuracy. Indeed, \cite{brogliato2002numerical} notes that the time-stepping scheme, ``is of order 1 therefore not very accurate unless $h$ is decreased a lot ... it should therefore be preferred for systems with a lot of events only''. However, in the relaxed formulation presented here the impact events are stretched over the duration of a full finite element (as in time-stepping schemes). As such this method maintains the ability to model systems with many near-simultaneous impacts, with the same loss of precision that time-stepping incurs in isolating the effects of the separate events. The main disadvantage of our method (like other contact-implicit methods) is that it is not suited for cases with a large number of interacting bodies as each possible interaction would require a dedicated set of complementarity constraints, thereby increasing the problem size considerably \cite{Posa2014AContact}.

One improvement of the presented method would be to use a more expressive control basis. In this formulation we have utilized piecewise-constant inputs (within the finite element) as a means to regularize and aid convergence of the singular control problems. Other input control profiles that vary over the finite element, but still provide the regularization to avoid the problems shown in Fig.~\ref{fig:hipTorqComp_fig}, would improve the performance, e.g.\ as in \cite{chen2016nested}. 

Another interesting future direction of this work would be to combine the variational methods described by \cite{Manchester2017} with orthogonal collocation. This can be achieved by discretizing the Lagrangian using Radau collocation, thereby increasing the accuracy while maintaining the attractive energy preservation properties of the variational methods \cite{marsden2001discrete}.

\section*{ACKNOWLEDGMENT}
The authors would like to thank Joseph Norby, Zachary Manchester, and Scott Kuindersma for their helpful discussions.

%%%%%%%%%%%%%%%%%%%%%%%%%%%%%%%%%%%%%%%%%%%%%%%%%%%%%%%%%%%%%%%%%%%%%%%%%%%%%%%%

%\addtolength{\textheight}{-11cm}   % This command serves to balance the column lengths

%%%%%%%%%%%%%%%%%%%%%%%%%%%%%%%%%%%%%%%%%%%%%%%%%%%%%%%%%%%%%%%%%%%%%%%%%%%%%%%%

%%%%%%%%%%%%%%%%%%%%%%%%%%%%%%%%%%%%%%%%%%%%%%%%%%%%%%%%%%%%%%%%%%%%%%%%%%%%%%%%

%%%%%%%%%%%%%%%%%%%%%%%%%%%%%%%%%%%%%%%%%%%%%%%%%%%%%%%%%%%%%%%%%%%%%%%%%%%%%%%%

%%%%%%%%%%%%%%%%%%%%%%%%%%%%%%%%%%%%%%%%%%%%%%%%%%%%%%%%%%%%%%%%%%%%%%%%%%%%%%%%
\bibliographystyle{IEEEtran}
\bibliography{roboticsRefs} % Entries are in the "refs.bib" file

\end{document}